\documentclass[10pt,twocolumn,letterpaper]{article}

\usepackage[pagenumbers]{cvpr} % To force page numbers, e.g. for an arXiv version

\usepackage{graphicx}
\usepackage{amsmath}
\usepackage{amssymb}
\usepackage{booktabs}

\usepackage{algorithm}
\usepackage{algorithmic}
\usepackage{xspace}

\usepackage{graphicx}
\usepackage{caption}
\usepackage{subcaption}
\usepackage{bm}
\usepackage{amsmath,amssymb}
\usepackage{booktabs}
\usepackage{multirow}
\usepackage[flushleft]{threeparttable}

\usepackage{multirow}
\usepackage{makecell}
\usepackage{multirow}
\usepackage{color}
\usepackage{wrapfig}

\usepackage[pagebackref,breaklinks,colorlinks]{hyperref}

\usepackage[capitalize]{cleveref}
\crefname{section}{Sec.}{Secs.}
\Crefname{section}{Section}{Sections}
\Crefname{table}{Table}{Tables}
\crefname{table}{Tab.}{Tabs.}

\def\cvprPaperID{1950} % *** Enter the CVPR Paper ID here
\def\confName{CVPR}
\def\confYear{2023}

\begin{document}

%%%%%%%%% TITLE - PLEASE UPDATE
\title{One Student Knows All Experts Know: From Sparse to Dense}

\author{Fuzhao Xue$^{1}$ \quad Xiaoxin He$^{1}$  \quad Xiaozhe Ren$^2$ \quad Yuxuan Lou$^1$ \quad Yang You$^1$\\
$^1$Department of Computer Science, National University of Singapore \\
$^2$Huawei Noah’s Ark Lab\\
}
\maketitle

%%%%%%%%% ABSTRACT
\begin{abstract}
Human education system trains one student by multiple experts. Mixture-of-experts (MoE) is a powerful sparse architecture including multiple experts. However, sparse MoE model is easy to overfit, hard to deploy, and not hardware-friendly for practitioners. In this work, inspired by the human education model, we propose a novel task, knowledge integration, to obtain a dense student model (OneS) as knowledgeable as one sparse MoE. We investigate this task by proposing a general training framework including knowledge gathering and knowledge distillation. Specifically, to gather key knowledge from different pre-trained experts, we first investigate four different possible knowledge gathering methods, \ie summation, averaging, Top-K Knowledge Gathering (Top-KG), and Singular Value Decomposition Knowledge Gathering (SVD-KG) proposed in this paper. We then refine the dense student model by knowledge distillation to offset the noise from gathering. On ImageNet, our OneS preserves $61.7\%$ benefits from MoE and achieves $78.4\%$ top-1 accuracy ImageNet with only $15$M parameters. On four natural language processing datasets, OneS obtains $88.2\%$ MoE benefits and outperforms the best baseline by $51.7\%$ using the same architecture and training data. In addition, compared with the MoE counterpart, OneS can achieve $3.7 \times$ inference speedup due to less computation and hardware-friendly architecture.
\end{abstract}

%%%%%%%%% BODY TEXT

\section{Introduction}
\label{Introduction}
%\section{Related Work}

%Motivation from Education
Revisiting how we become a researcher, most people learn from multiple teachers (\ie experts). Existing work~\cite{bransford1999people} in education also shows that experts from different subjects can help students reach deep understanding and train more talents. The students who integrate knowledge from experts can become as knowledgeable as the set of these experts fast. Inspired by such human education model, this work focuses on training a powerful deep learning model by collecting knowledge from a set of experts.

%For researchers, they learn various courses from professors at school and then can be prepared well for research. Motivated by the human education pattern, this work devotes to train a powerful deep learning model by collecting knowledge from a set of teachers (\ie experts). 

%\begin{wrapfigure}{r}{0.4\textwidth}
\begin{figure}[t]
%\vspace{-0.3cm}
\begin{center}
\centerline{\includegraphics[width=0.4\textwidth]{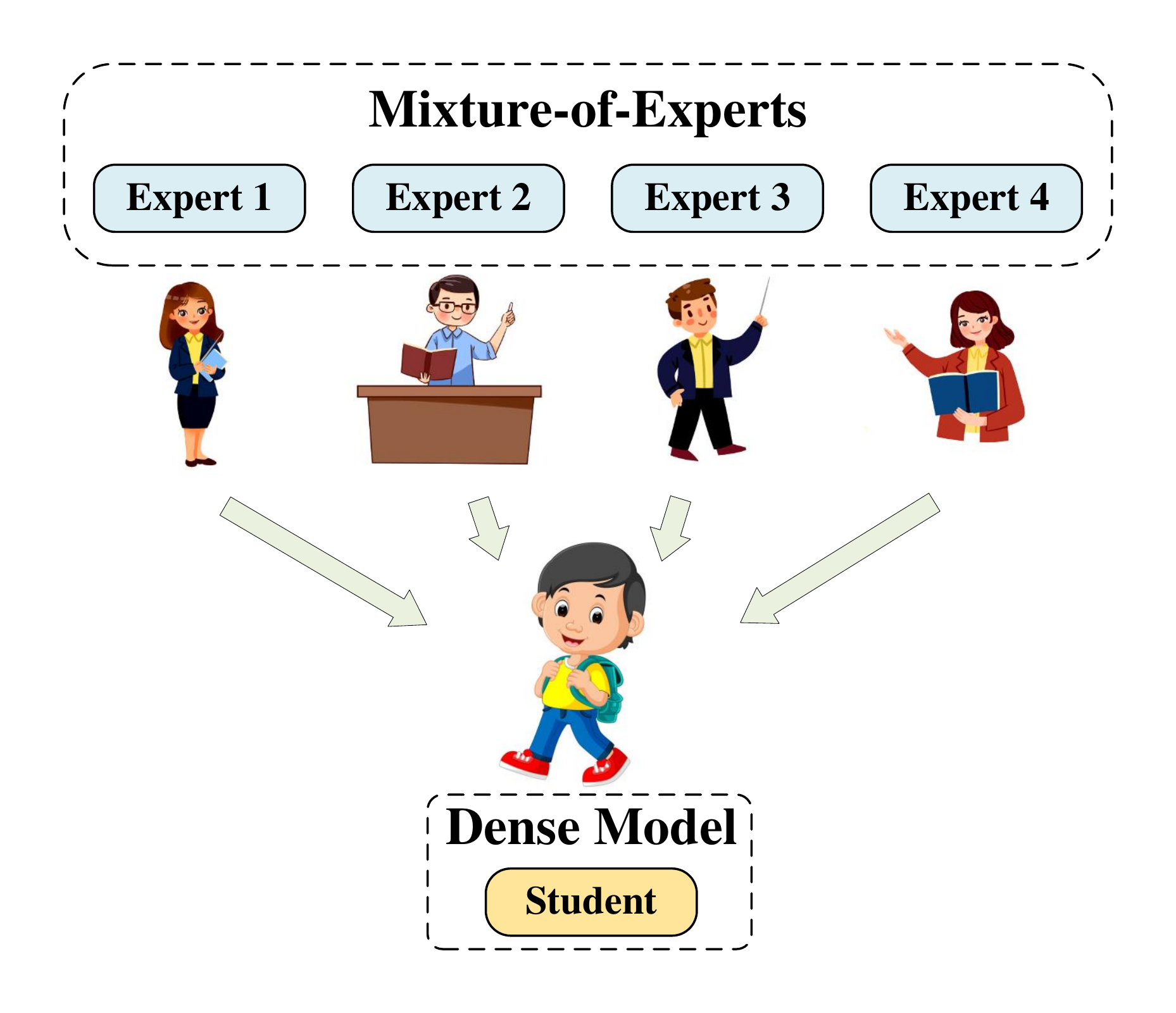}}
\vskip -0.2in
\caption{Human education model matches MoE and dense model.}
\label{intro:motivation}
\end{center}
\vspace{-1.2cm}
\end{figure}
%\end{wrapfigure}

Recent study in deep learning proposed mixture-of-experts (MoE), a deep neural network with multiple experts. Each expert is a sub-neural network in the whole model. The key idea of MoE is to divide and conquer the task. MoE encourages each expert to learn from a task-specific subset of the input. For each subset of the input, there would be only a sub-network activated. Such sparse computation of MoE enables us to scale model to trillions of parameters with comparable computation cost~\cite{fedus2021switch}.

%Unfortunately, expert parallelism is communication expensive and hard to implement. For GPU clusters, all-to-all operation is too slow to scale the MoE model up. 

The MoE model is powerful and achieved promising results due to its large but sparse-activated model capacity. However, MoE is easy to overfit. We usually pre-train an MoE on a large dataset and then fine-tune it on various downstream tasks. In most cases, these downstream tasks are the target problem we want to solve. Compared with dense models, more trainable parameters and sparse conditional computation introduce overfitting~\cite{Xue2021GoWI,lou2021sparse} during fine-tuning, especially when the scale of dataset is not large enough. In addition, even if we trained an MoE model successfully, it is hard to deploy. For MoE with trillions of parameters, we need to deploy different experts on different devices to reduce the memory consumption on device (\eg GPU, TPU). Third, MoE model is not hardware-friendly. Expert parallelism is communication expensive. For GPU clusters, the all-to-all operation is too slow to scale the MoE model up. Besides, the gating function includes numerous operations to create token masks, select top-k experts, and perform cumulative-sum to find the token-id going to each expert and sparse matrix-multiply~\cite{rajbhandari2022deepspeed}. All these operations are wasteful due to the sparse tensor representation. More importantly, they are extremely slow due to many kernel call invocations. In summary, the sparse MoE model is powerful, but it is relatively hard to use in practice. The dense model is widely used but weaker than the sparse model with comparable computation cost. Then, is it possible to combine the strength of sparse and dense model to train a model that is both effective and easy to use?

In this work, inspired by human education model, we propose a new task, \ie knowledge integration. As a general training framework, knowledge integration includes two steps, \ie knowledge gathering and knowledge distillation. In knowledge gathering, we treat each expert in MoE as a specialist in human education. The student is a dense model, and we are to collect knowledge from all experts and assign the knowledge to the student. To gather knowledge from experts, as the first work focusing on this task, we investigate four different possible solutions, \ie summation, averaging, Top-K Knowledge Gathering (Top-KG), and Singular Value Decomposition Knowledge Gathering (SVD-KG) proposed in this work. For the Top-KG and SVD-KG, we use Top-K selection or SVD to extract key knowledge from different experts of a pre-trained MoE, and then, we initialize the feed-forward network (FFN) layers for a dense model to approximate the MoE. To further refine the model from noise, we use knowledge distillation~\cite{hinton2015distilling} to fine-tune the student. Please note in knowledge distillation stage, we use the whole MoE model to teach the student dense model. The final student model has the same architecture as a standard dense model, but, it would cover the knowledge of MoE with many experts and much more trainable parameters. The framework described above matches well with the human education model, one student integrates knowledge from multiple experts so that the student can learn fast. 

%would cover the 

Our contributions are summarized as follows: 

\begin{itemize}

\item We propose a new task, knowledge integration. The goal is to combine the effectiveness of the sparse MoE model and the usability of the dense model. To our best knowledge, this is the first work focusing on learning a dense model from a pre-trained MoE model. 

\item We propose to solve knowledge integration in two steps, knowledge gathering and knowledge distillation. To gather, we first investigate four different possible knowledge gathering methods, \ie summation, averaging, Top-KG and SVD-KG proposed in this paper. Top-KG and SVD-KG are novel methods to extract and merge key knowledge from experts of a pre-trained MoE to initialize a dense model.

\item We evaluate our general training framework in different areas, \ie computer vision and natural language processing. On ImageNet, compared with baselines, our OneS preserve $23.1\%$ more benefits from MoE. On natural language processing benchmarks, we achieve $88.2\%$ MoE benefits with only $46\%$ parameters, and we outperform baselines (\eg Distill, Switch) using almost the same architecture and training data. Also, due to the hardware-friendly model architecture, OneS can achieve $3.7 \times$ inference speedup over the MoE counterpart.

\end{itemize}

%\begin{itemize}

%\item We propose a new task, knowledge integration. The goal is to combine the effectiveness of Sparse MoE model and the usability of dense model. To our best knowledge, this is the first work focusing on learning a dense model from a pre-trained MoE model. 

%\item We propose to solve knowledge integration in two steps, knowledge gathering and knowledge distillation. To gather, we propose Singular Value Decomposition Knowledge Gathering, a new approach to extract key knowledge from experts of a pre-trained MoE and initialize a dense model.

%\item  

%\end{itemize}

\section{Preliminary}
\label{Preliminary}

\subsection{Mixture-of-Experts}
\label{Preliminary:Mixture-of-Experts}

Mixture-of-experts is a typical conditional computation model. In this work, we use a pre-trained MoE model as a teacher, and a dense model as a student to imitate the human education model. Therefore, we briefly review MoE first. Given one MoE model with $E$ trainable experts and input representation $x\in \mathbb{R}^D$, the output of MoE model can be formulated as ~\cite{shazeer2017outrageously},:
\begin{equation}\label{eq:MoE}
\mathrm{MoE}(x)=\sum_{i=1}^E {G(x)}_i {e_i(x)}
\end{equation}
where ${e_i(\cdot)}$ is a non-linear transformation $\mathbb{R}^D \to \mathbb{R}^D$ of the $i^{\mathrm{th}}$ expert, and ${G(\cdot)}:\mathbb{R}^D \to \mathbb{R}^E$ is the gating network, ${G(x)}_i$ is the routing weights of $x$ to the $i\text{-}th$ expert. Usually, both $e(\cdot)$ and $G(\cdot)$ are parameterized by neural networks. Please note the output of $G(\cdot)$ should be activated by softmax function:
\begin{equation}\label{eq:topK}
\mathrm{G}(x)=\mathrm{topK}(\omega(h(x)+\epsilon))
\end{equation}
where $\omega$ is the softmax function, $h(\cdot)$ is a linear layer mapping $\mathbb{R}^D \to \mathbb{R}^E$, and $\epsilon \sim \mathcal{N}(0,\frac{1}{E^2})$ is a Gaussian noise for exploration of expert routing. The top-K selection is a key module to activate sub-network sparsely. We usually set $K$ as 1 or 2 for comparable computation cost with the corresponding dense model.

When training MoE model, if we have no regularization, most tokens may be dispatched to a small portion of experts, and other experts receive few tokens. Such an imbalanced assignment would lead to lower efficiency and inferior accuracy~\cite{lepikhin2020gshard, fedus2021switch}. Therefore, to achieve balanced workload for different experts, we usually combines router $g(\cdot)$ with load balance loss~\cite{lepikhin2020gshard} $\mathrm{L_{balance}}$:

\begin{equation}\label{eq:balance_loss}
 \mathrm{L_{balance}} = E \cdot \sum_{i=1}^E m_i \cdot P_i
\end{equation}
where $m$ is a vector and the $i^{\mathrm{th}}$ element of $m$ represents the fraction of tokens dispatched to expert $i$:
\begin{equation}
m_i = \frac{1}{N} \sum_{j=1}^{N} \mathrm{k}(x_j)_i
\end{equation}
where $N$ is the number of tokens to route, $\mathrm{k}(x_j)$ is an index vector from top-K function. Since the index vector generation here is non-differentiable, we define $P_i$ as:
\begin{equation}
P_i = \omega(h(x)+\epsilon)_i
\end{equation}
where $P$ is $g(x)$ without top-K routing. When we minimize $\mathrm{L_{balance}}$, we can see both $m$ and $P$ would close to a uniform distribution. 

%so that model can achieve higher efficiency and effectiveness. 

The trainable router here can also be replaced by non-trainable modules, \eg BASE layer~\cite{Lewis2021BASELS}. This work focuses on integrating knowledge from a pre-trained MoE instead of MoE variants.

%According to the formulation above, when $g(\cdot)$ is a sparse vector, only part of experts would be activated and updated by back-propagation during training. In this paper, for both vanilla MoE and our WideNet, each expert is an FFN layer.

%\revise{FUZHAO: TODO}

\begin{figure*}[t]
%\vspace{-0.5cm}
\begin{center}
\centerline{\includegraphics[width=0.75\textwidth]{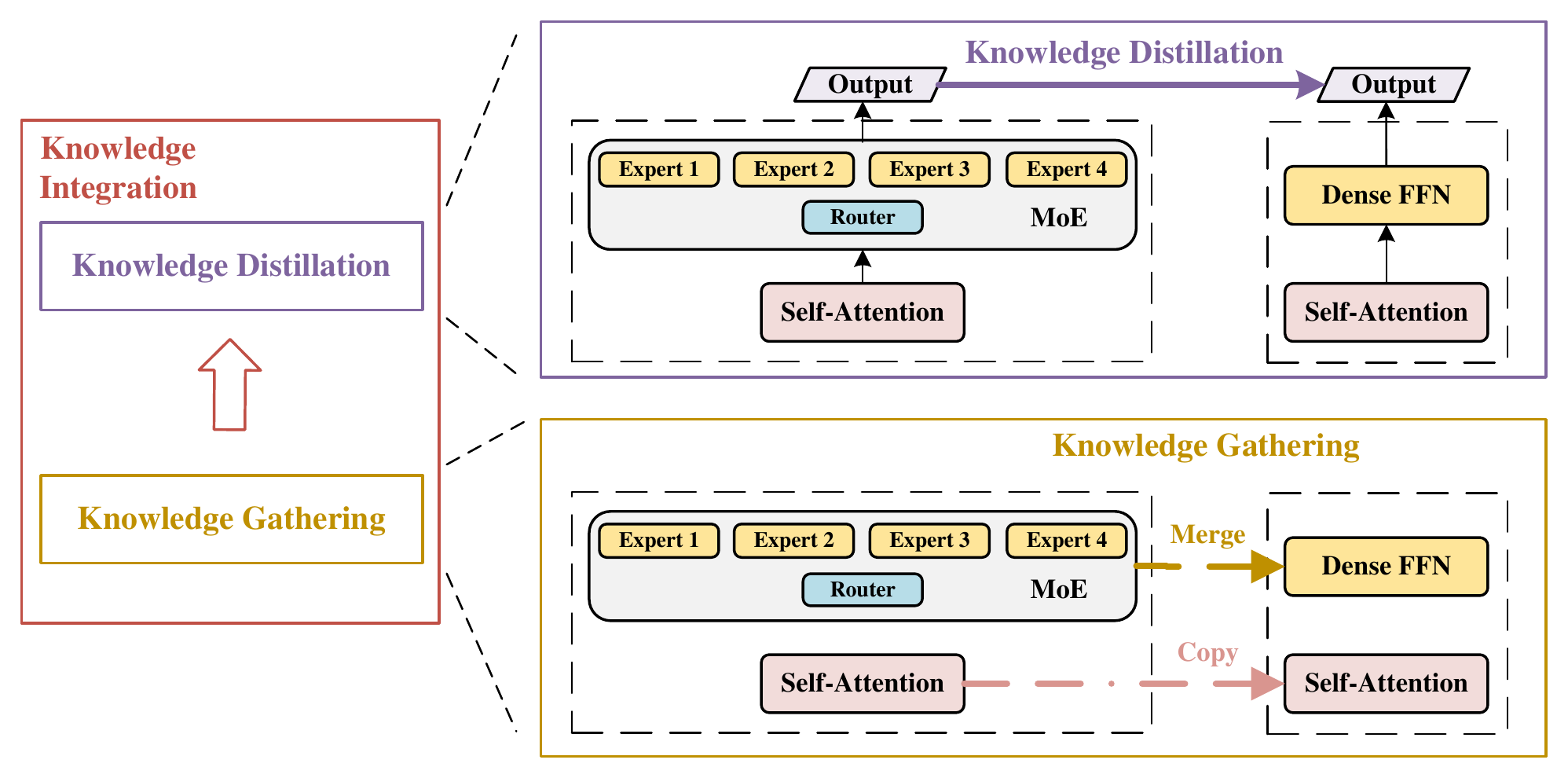}}
\caption{An overview of our general training framework proposed. The overall training framework is knowledge integration, and it includes two stages, knowledge gathering and knowledge distillation. In knowledge gathering, we investigate four different methods to merge the knowledge from MoE, including summation, averaging and Top-KG and SVD-KG.}
\label{fig:overview}
\end{center}
\vspace{-0.8cm}
\end{figure*}

\subsection{Problem Formulation}
\label{Preliminary:Problem_Formulation}
We have two stages in the knowledge integration framework proposed in this work: (1) knowledge gathering from MoE; (2) knowledge distillation to further refine the new dense model (\ie student). For the first stage, given $E$ experts $\{e_1(\cdot),e_2(\cdot),\dots,e_E(\cdot)\}$, we are to maximize the knowledge covered in the dense model $s(\cdot)$. We use transformer-based MoE to introduce our framework due to its popularity. Given input representation $x$, within one transformer block, each expert is an FFN, which can be formulated as:

\begin{equation} \label{eq:expert_form}
e_i(x)=f^{i}_{2}(\sigma(f^{i}_1(x)))
\end{equation}
where $f^{i}_1(\cdot)$ and $f^{i}_2(\cdot)$ and linear transformations of $i^{th}$ expert, $\sigma(\cdot)$ is the activation functions. For the dense student, we have the same architecture but different trainable parameters:

\begin{equation} \label{eq:student form}
s(x)=g_{2}(\sigma(g_1(x)))
\end{equation}
where $\sigma(\cdot)$ would be the same activation function as experts. The only difference is the trainable parameters in linear transformations. Then, our target is to approximate the trainable parameters of $g_1$ and $g_2$ according to $\{f^{1}_1,\dots, f^{E}_1\}$ and $\{f^{1}_2,\dots, f^{E}_2\}$, respectively. We define this target as knowledge gathering from MoE.

The second stage is fine-tuning the dense student to minimize the difference between teacher output and student output. We can easily find this task closer to knowledge distillation~\cite{hinton2015distilling}, so in this paper, we follow the typical KD approaches as our solution. 

Our goal is to preserve MoE's benefits by a dense student as much as possible. So, we define a metric, MoE benefits, to measure the ability of a dense student to integrate knowledge from the MoE counterpart. The MoE benefits can be written as:
\begin{equation} \label{eq:moe_benefits}
\mathrm{MoE~benefits}= \frac{\mathrm{score}_{student}-\mathrm{score}_{dense}}{\mathrm{score}_{MoE}-\mathrm{score}_{dense}}
\end{equation}
where $\mathrm{score}$ can be any metric to evaluate the model. For instance, $\mathrm{score}$ is accuracy for image classification. The $\mathrm{score}_{dense}$ here denotes the dense model's performance without knowledge integration proposed.

%As the second problem has good existing solution, we devote to solve the first one in this work. 

%\subsection{WideNet}

\section{Approach}
\label{Approach}

In general, the final target of this work is to obtain a dense student model that is easy to use and as effective as the sparse MoE. To this end, we propose a general training framework, knowledge integration, to integrate knowledge from sparse MoE teacher to dense student. The proposed knowledge integration includes two stages: knowledge integration from MoE and knowledge distillation to refine the student. An overview of the proposed general training framework is shown in Figure~\ref{fig:overview}. The first step is to initialize the dense student. For most trainable layers (\eg embedding layer, attention layer, normalization layer), the teacher and the student have the same structure (We name such layers as perfectly matched layers in this work.), so we can copy the weights from teachers following Switch Transformer~\cite{fedus2021switch} directly. The challenging part is the MoE layer. MoE layer has much more trainable parameters than the dense counterpart with a single FFN layer, and each expert is actually an FFN layer with unique weights and biases. The core issue is to incorporate knowledge from different FFN experts and assign the knowledge to one single FFN in the student model. To this end, we investigate four different possible knowledge gathering methods, i.e., summation, averaging, Top-KG and SVD-KG. Then, knowledge distillation is to fine-tune the initialized model to further improve performance. 

\subsection{Knowledge Gathering from MoE}

We first formulate our KG task. Given an MoE layer with $E$ experts, the target here is to gather knowledge from all experts for one dense student. According to Eq.~\ref{eq:expert_form} and Eq.~\ref{eq:student form}, each expert comprises two linear layers, and the student shares the same model structure with one single expert. For brevity, we treat each expert as one linear transformation to show our idea, which can be expanded to multiple linear layers easily. For E linear layers $\{f^1, f^2, \dots, f^E \}$, each linear layer $f^i(\cdot):\mathbb{R}^{d_1} \to \mathbb{R}^{d_2}$ with weights $W_f^i \in \mathbb{R}^{d_{1} \times d_{2}}$ and bias $b_f^i \in \mathbb{R}^{d_2}$,
\begin{equation}
\begin{aligned}
 & \mathrm{KG}(f^1, f^2, \dots, f^E) \\
= & \mathrm{KG}(W_f^1, W_f^2, \dots, W_f^E;b_f^1, b_f^2, \dots, b_f^E)  \\
\approx &(W_g ; b_g) = g \\
\end{aligned}
\end{equation}
where $g(\cdot):\mathbb{R}^{d_1} \to \mathbb{R}^{d_2}$ is a linear layer with $W_g \in \mathbb{R}^{d_{1} \times d_{2}}$ and bias $b_g \in \mathbb{R}^{d_2}$.

%\subsubsection{Summation and Averaging}

Before merging the weights, we first initialize $b_g$ from different experts. Since it has much fewer trainable parameters, we simply average the bias vector from different experts:
\begin{equation}
b_g=\frac{1}{E} \sum_{i=1}^E b_f^i
\end{equation}
We employ such a simple policy because knowledge stored in bias is much less than in weights, due to fewer trainable parameters. We justify this assumption by experiments in Appendix~\ref{Appendix:less-know-bias}.

After copying the weights and bias in the perfectly matched layers and averaging bias in the MoE layers, we initialize the dense student model weights by sparse MoE. As the first work focusing on this task, we investigate four methods to gather the knowledge, \ie summation, averaging, Top-KG and SVD-KG. The first two are the most straightforward methods. We also propose two novel approaches, Top-KG and SVD-KG to extract key knowledge from different experts of a pre-trained MoE.  
 %There are multiple experts in one MoE layer. All these experts have the same shape of trainable matrices. That is, the trainable matrix of student dense model can be perfectly matched to only one expert. The challenge here is, if we simply combine all experts by averaging the matrices, it is unavoidable to induce much noise. Therefore, 

\subsubsection{Summation and Averaging}

For weights in MoE, we first consider two simple methods. The first one is the summation:
\begin{equation}
W_g=\sum_{i=1}^E W_f^i
\end{equation}
and the second one is averaging:
\begin{equation}
W_g=\frac{1}{E} \sum_{i=1}^E W_f^i
\end{equation}

Although these two gathering methods are simple, as the first work focusing on this task, we investigate them to pave the way for gathering knowledge from MoE models.

\subsection{Top-K Knowledge Gathering}

We also propose two novel methods to gather knowledge. For weights, in MoE, a wide over-parameterized model with much more trainable parameters, it is challenging to cover all knowledge in a narrow dense model. Therefore, we have to extract the key knowledge from each expert and then merge them into a single small dense model. Then, the question is, how can we extract the key knowledge of each trainable matrix (\ie weights)? We first propose Top-K knowledge gathering to extract the sub-matrix of each expert. For $i^{\mathrm{th}}$ expert weight matrix $W^{\mathrm{i}} \in \mathbb{R}^{d_{1} \times d_{2}}$, we calculate the l2 norm of each column as $l^{\mathrm{i}} \in \mathbb{R}^{d_{1}}$. We then use Top-K selection to pick $K$ columns of $W^{\mathrm{i}}$ according to $l^{\mathrm{i}}$, where $K=\frac{d_{2}}{E}$. The extracted matrix $W^{\mathrm{i}}_g \in \mathbb{R}^{d_{1} \times K}$. Then we concatenate the extracted matrices from all experts as final student initialization $W_g \in \mathbb{R}^{d_{1} \times d_{2}}$.

In practice, since each expert has two linear layers $W^{\mathrm{i}_1} \in \mathbb{R}^{d_{1} \times d_{2}}$ and $W^{\mathrm{i}_2}  \in \mathbb{R}^{d_{2} \times d_{1}}$, there would be a column-mismatch for two extracted matrices from the same expert if we select the sub-matrices of these two matrices independently. To alleviate this issue, we calculate the l2 norm of each column in $W^{\mathrm{i}_1}$ and the l2 norm of each row in $W^{\mathrm{i}_2}$. The sum of these two l2 norm vectors, \ie $l^{\mathrm{i}} \in \mathbb{R}^{d_{1}}$ is fed into Top-K selection and then extract the sub-matrix.

\subsubsection{SVD Knowledge Gathering}

%Given $E$ experts, we assume $i\text{-}th$ expert $e_i$ is a linear layer $f^i(\cdot):\mathbb{R}^{d_1} \to \mathbb{R}^{d_2}$ with weights $W_f^i \in \mathbb{R}^{d_{1} \times d_{2}}$ and bias $b_f^i \in \mathbb{R}^{d_2}$. Then, the dense student $s$ is another linear layer $g(\cdot):\mathbb{R}^{d_1} \to \mathbb{R}^{d_2}$ with $W_g \in \mathbb{R}^{d_{1} \times d_{2}}$ and bias $b_g \in \mathbb{R}^{d_2}$ The SVD-KG proposed in this work is to initialize $W_g$ and $b_g$ and make $s(x)$ closer to $\mathrm{MoE}(x)$.
We investigate another novel way to extract key knowledge from experts. Low-rank compression~\cite{chen2021drone} has shown promising results in capturing key knowledge, which was used to convert a not low-rank matrix to a rank-$k$ decomposition of the weight matrix. Such a low-rank matrix can approximate the knowledge of the whole matrix. On this basis, we can merge the low-rank matrix easier by reconstructing a high-rank matrix from multiple low-rank matrices. Please note, in this work, obtaining rank-$k$ decomposition is not our target. Instead, the rank-$k$ decomposition is just an intermediate step of our decomposing and merging. In this work, we propose to use SVD to extract key knowledge and merge them to initialize another dense matrix:
\begin{equation}
W_f^i= U_f^i S_f^i {V_f^i}^T\approx {U_f^i}_{K^i} {S_f^i}_{K^i} {V_f^i}_{K^i}^T
\end{equation}
where $U_f^i \in \mathbb{R}^{d_1 \times d_1}$ and $V_f^i \in \mathbb{R}^{d_2 \times d_2}$ are unitary matrices, $S_f^i \in \mathbb{R}^{d_1 \times d_2}$ is a diagonal matrix. We usually select the top-K elements in $S_f^i$ and then construct ${U_f^i}_{K^i} \in \mathbb{R}^{d_1 \times K^i}$, ${S_f^i}_{K^i} \in \mathbb{R}^{{K^i} \times {K^i}}$ and ${V_f^i}_{K^i} \in \mathbb{R}^{d_2 \times {K^i}}$ to approximate $W_f^i$.

When $k$ is fixed, every matrix has the rank-$k$ decomposition to approximate the original matrix. However, we cannot guarantee the key knowledge in every expert can be covered by a fixed rank-$k$ decomposition. Thus, we define an adaptive SVD ratio $\lambda \in (0,1]$ to ensure:
\begin{equation}\label{eq:auto_SVD}
 \rho({S_f^i}_{K^i}) \approx \lambda \rho(S_f^i) 
\end{equation}
where $\rho(S_f^i)$ denotes the sum of diagonal elements of $S_f^i$. If $\lambda=1$, all ranks would be preserved for a full-rank matrix. We then collect the decomposition of each expert and concatenate them as:
\begin{equation}
\small
\begin{aligned}
[~U_g~] &= \left[ \begin{array}{ccc}
{U_f^1}_{K^1} & \dots & {U_f^E}_{K^E} 
\end{array} 
\right ], \\
[~S_g~] &= \left[ \begin{array}{ccc}
{S_f^1}_{K^1} & ~ & ~ \\
~ & \ddots & ~ \\
~ & ~ & {S_f^E}_{K^E}
\end{array} 
\right ], \\
[~V_g~] &= \left[ \begin{array}{c}
{V_f^1}_{K^1} \\ 
\vdots \\
{V_f^E}_{K^E} 
\end{array} 
\right ] \\
\end{aligned}
\end{equation}

We can then obtain $W_g$ as:
\begin{equation}\label{eq:assign_stu}
W_g= U_g S_g {V_g}^T
\end{equation}
$W_g$ is a rank-$K_g$ matrix, where $K_g=\Sigma_{i=1}^E K^i$, covering the key knowledge of every expert.

 % but we assume the knowledge from experts is totally independent That is, we assume knowledge from other experts has no impact on the left expert. 
After SVD-KG, knowledge has been integrated from pre-trained MoE. However, during knowledge gathering, it is unavoidable to induce noise when we remove conditional computation. Detailed analysis of the induced noise during gathering can be found in Appendx~\ref{appendix:noise}.

%\begin{theorem}
%\label{thm:1}
%Assume each expert has one linear layer without bias. Given $E$ rank-$k$ matrices $W_f$ from $\mathrm{MoE}(\cdot)$, we have that one rank-$Ek$ matrix $W_s$ can approximate the MoE:
%\begin{equation}\label{eq:SVD}
%\begin{aligned}
%\mathrm{MoE}(x) &\approx \sum_{i=1}^E {U_f^i}_{Ek} {S_f^i}_{Ek} {V_f^i}_{Ek}^T x \\
%&= {U_f^i}_{Ek} {S_f^i}_{Ek} {V_f^i}_{Ek}^T \\
%&= s(x)
%\end{aligned}
%\end{equation}
%\end{theorem}

%The proof of Theorem~\ref{thm:1} is provided in \revise{Appendix}. By Eq.~\ref{eq:SVD}, we can initialize the weights in dense model integrating most knowledge from all experts in MoE.

 %The FLOPs here means the floating point operations in FFN layer or MoE layer. We only report the FLOPs in FFN or MoE because FLOPs at other layers are totally same.

\subsection{Knowledge Distillation}\label{sec:KD}

To mine the knowledge from noise, we adopt soft knowledge distillation~\cite{hinton2015distilling} to fine-tune the dense student. Soft distillation minimizes the Kullback-Leibler divergence between the output of the teacher and the student. The corresponding distillation loss can be written as:
\begin{equation}
\mathrm{L}_{distill}^{soft} = T^2 \mathrm{L_{KL}}(\omega(z_{s}/T), \omega(z_{t}/T))
\end{equation}
where $\omega$ is the softmax function, $L_{KL}$ is Kullback-Leibler divergence loss, $z_s$ and $z_t$ are the logits of student and teacher, respectively, and $T$ is the softmax temperature. We also considered hard-label distillation~\cite{pmlr-v139-touvron21a} and compared its performance with soft distillation. Please see Appendix~\ref{Appendix:hard-label} for details.

\subsection{Optimization}

Our final loss function is simple:
\begin{equation}\label{eq:loss}
\mathrm{L_{total}} = \alpha \mathrm{L_{main}} + (1-\alpha) \mathrm{L_{distill}}
\end{equation}
where $\alpha$ is used to balance the main loss and the distillation loss. The main loss depends on the task. For instance, to classify images, it is cross-entropy. For BERT pre-training, it should be the masked language modeling loss and next sentence prediction loss. The distillation loss here can be either soft distillation loss or hard-label distillation loss. Since our pre-trained MoE is fixed during knowledge distillation, we do not need the load balance loss of MoE-based transformer.

%For hard-label distillation, $\alpha$ is usually set as $0.5$. 
%In Section~\ref{Experiments:CV:Ablation}, we found there is no significant difference in performance when using different types of distillation loss. Therefore, we adapt soft distillation, a more widely used approach as our default choice. Since our pre-trained MoE is fixed during knowledge distillation, we do not need the load balance loss of MoE.

\section{Experiments}
\label{Experiments}

\begin{table*}[t]
%\begin{wraptable}{r}{0.7\textwidth}
%\vspace{-0.5cm}

%\vspace{-0.3cm}
\begin{center}
%\begin{sc}
\begin{small}
\begin{tabular}{l l l l l l l l}
\toprule
~ & Model    & MoE or Dense & Para Sharing  & \#Para & ImageNet & Benefits(\%) \\ \midrule
\multirow{4}{*}{ViT} & ViT-B   & Dense &             & 87M       & 78.6    & -    \\
& ViT-L  & Dense &      & 305M         & 77.5 & -  \\
 & ViT-B   & Dense &    \checkmark         & 10M       & 72.8   & -    \\
& ViT-L  & Dense &  \checkmark    & 15M         & 76.9 & -  \\ \midrule
\multirow{2}{*}{Teacher} & WideNet-B  & MoE &  \checkmark       & 29M        & 77.5   & -    \\
& WideNet-L   & MoE &  \checkmark         & 40M     & 79.5   & -    \\ \midrule
\multirow{4}{*}{Baseline} & Distill-B   & Dense &  \checkmark       & 10M        & 73.8   & 21.3   \\
& Distill-L  & Dense &  \checkmark     & 15M     & 77.3   & 15.3  \\
& Switch-B   & Dense &  \checkmark     & 10M        & 74.8  & 42.6     \\
& Switch-L    & Dense &  \checkmark      & 15M     & 77.8   & 34.6    \\ \midrule
\multirow{6}{*}{Ours} & OneS-B Sum  & Dense &  \checkmark        & 10M        & 75.2   & 51.1   \\
& OneS-L Sum     & Dense &  \checkmark       & 15M     & 78.2   &  48.1   \\
 & OneS-B Avg  & Dense &  \checkmark        & 10M        & 75.3   & 53.2   \\ 
 & OneS-L Avg     & Dense &  \checkmark       & 15M     & 78.0  &  40.7   \\
  & OneS-B Top-K  & Dense &  \checkmark        & 10M        & 75.3   & 53.2   \\ 
 & OneS-L Top-K     & Dense &  \checkmark       & 15M    & \textbf{78.4}   &  \textbf{57.7}   \\
 & OneS-B SVD   & Dense &  \checkmark        & 10M        & \textbf{75.7}   & \textbf{61.7}   \\ 
& OneS-L SVD     & Dense &  \checkmark       & 15M     & \textbf{78.4}   &  \textbf{57.7}   \\
\bottomrule
\end{tabular}
\end{small}
%\end{sc}
\end{center}
\vspace{-0.3cm}
\caption{Top-1 Accuracy and MoE Benefits(\%) on ImageNet pre-training. As we defined in Eq.~\ref{eq:moe_benefits}, MoE Benefits denotes the percentage of performance improvement from MoE that can be preserved in the dense student model. The Para Sharing denotes whether the trainable parameters are shared across transformer blocks. We use such model (\ie WideNet) as the MoE layer dominates the trainable parameters, which can verify the effectiveness of knowledge integration methods directly. For the ViT without parameter sharing, we can usually observe the overfitting issue when training on ImageNet.}
\label{cv:table:main}
\vspace{-0.3cm}
\end{table*}

\begin{table}[t]
%\begin{wraptable}{r}{0.45\textwidth}
%\vspace{-0.8cm}

%\vspace{-0.3cm}
\begin{center}
%\begin{sc}
\begin{small}
\begin{tabular}{l l l l}
\toprule
~ & Model     & \#Para   & Cifar10 \\ \midrule
\multirow{2}{*}{ViT} & ViT-B               & 85M       &  98.3      \\
& ViT-L              & 305M         &   98.2    \\ \midrule
\multirow{2}{*}{Teacher} & WideNet-B         & 27M        &     98.4    \\
& WideNet-L          & 38M     &       98.8    \\ \midrule
\multirow{2}{*}{Baseline} & Switch-B         & 9M        &     97.9    \\
& Switch-L          & 13M     &       98.3  \\ \midrule
\multirow{2}{*}{Ours} & OneS-B         & 9M        &    98.1    \\
& OneS-L          & 13M     &       \textbf{98.5}  \\
\bottomrule
\end{tabular}
%\end{sc}
\end{small}
\end{center}
\vspace{-0.3cm}
\caption{Top-1 Accuracy on Cifar10 fine-tuning. We use our default knowledge gathering choice, SVD-KG, to gather the knowledge during pre-training. That is, for OneS, we finetune a dense model without knowledge distillation. }
\label{cv:table:cifar}
%\vspace{-0.6cm}
\end{table}
%~\cite{dosovitskiy2020image}
%~\cite{Xue2021GoWI}
%~\cite{artetxe2021efficient}
%~\cite{fedus2021switch}

\subsection{Computer Vision}
\label{Experiments:CV}

%\subsubsection{Experimental Settings}
%\label{Experiments:CV:setting}

\textbf{Experimental settings} To evaluate our general training framework, we conduct experiments on two different areas, computer vision and natural language processing. \textbf{Datasets} For vision, we select two widely-used image classification benchmarks, ILSVRC-2012 ImageNet~\cite{deng2009imagenet} and Cifar10~\cite{krizhevsky2009learning}, as platforms to evaluate our framework on computer vision. ILSVRC-2012 ImageNet dataset we used in this work has 1k classes and 1.3M images. We denote it as ImageNet in the following experiments for brevity. \textbf{Baselines} As we are the first work, to our best knowledge, focusing on integrating knowledge from a pre-trained MoE, the only two existing strong baselines are the knowledge distillation framework proposed in Meta AI MoE~\cite{artetxe2021efficient} and Switch Transformer~\cite{fedus2021switch}. The first one simply initializes the student dense model randomly. The second work initializes the dense model with the non-expert weights. That is, they simply copy the layer which can be perfectly matched into the dense model. For the weights that cannot be matched (\ie experts), they skip the initialization from MoE and train these layers from scratch instead. In our work, for brevity, we denote these two approaches as Distill and Switch, respectively. We also report the result of Vision Transformer (ViT) on the same setting to compare the parameter efficiency. \textbf{Teacher} In our training framework, we need an MoE model to initialize our dense student model (\ie knowledge gathering) and perform knowledge distillation. In this work, we apply the pre-trained WideNet~\cite{Xue2021GoWI}\footnote{We try two different scales of WideNet (\ie WideNet-Base, WideNet-Large) as our teacher, respectively. } as the platform. WideNet is an MoE-based transformer with only one trainable transformer block. This transformer block uses MoE instead of FFN layer to learn the local representation. The main focus of this paper is to verify the knowledge in the pre-trained MoE can be preserved in the dense student, so we use WideNet as our teacher model to verify the effectiveness of our approach in a more straightforward manner. \textbf{Hyper-parameters} For a fair comparison, we follow the data augmentation used in teacher model: Inception-style pre-processing, Mixup~\cite{zhang2017mixup}, RandAugment~\cite{cubuk2020randaugment} and label smoothing~\cite{szegedy2016rethinking,yuan2020revisiting}. We use LAMB~\cite{you2019large} optimizer. Batch size and learning rate are set as 4096 and 0.004, respectively. For the teacher model, all settings of WideNet~\cite{Xue2021GoWI} are the same as reported in their paper. Please note we freeze all trainable weights of the teacher model (\ie WideNet) in the knowledge distillation stage of OneS. For distillation hyper-parameters, we set $\alpha$ as 0.25 and temperature $T$ as 1.0. Linear learning rate decay is applied. 

We also fine-tune our pre-trained student model on Cifar-10. The setting is the same as ViT and WideNet. We use SGD optimizer with momentum. Following existing works, label smoothing and warm-up are removed. Please see Appendix for other training details.

\begin{table*}[t]
%\vspace{-0.5cm}

%\vspace{-0.3cm}
\begin{center}
%\begin{sc}
\begin{small}
\resizebox{1.0\textwidth}{!}{
\begin{tabular}{l l lllllll ll}
\toprule
& Model                   & \#para & FLOPs & Speedup & SQuAD1.1 & SQuAD2.0 & MNLI & SST-2  & Avg & Benefits(\%)\\ \midrule
\multirow{1}{*}{Teacher} & WideNet   & 26M & $2.4 \times$ & $1.0 \times$  & 89.6/82.7 & 80.6/77.4 &  82.6 & 91.1  & 84.71 & - \\ \midrule
%& WideNet 8 experts  & 45M & $2.4 \times$   & 90.0/82.7 & 80.6/77.7 & 83.3 & 91.9 &  85.2 & \\ 
%& WideNet 16 experts & 83M & $2.4 \times$   & 90.9/83.8 & 81.0/77.9 & 84.1 & 92.2  & 85.8 & \\ \midrule
\multirow{3}{*}{Baseline} & ALBERT                & 12M & $1.0 \times$ & $3.7 \times$  & 89.3/82.3 & 80.0/77.1 & 81.5 & 90.3 & 84.03 & 0.0 \\
& Distill  & 12M & $1.0 \times$ & $3.7 \times$ & 89.4/82.7 & 79.8/76.6 & 81.9 & 90.7  & 84.21 & 26.5 \\
& Switch  & 12M & $1.0 \times$ & $3.7 \times$ & 89.5/82.6 & 79.9/77.0 & 82.0 & 90.3  & 84.20 & 25.0 \\ \midrule
Ours & OneS  & 12M & $1.0 \times$ & $3.7 \times$ & \textbf{89.7/83.0} & \textbf{80.2/77.1} & \textbf{82.3} & \textbf{91.2}  & \textbf{84.63} & \textbf{88.2} \\ 
\bottomrule
\end{tabular}
}

%\end{sc}
\end{small}
\end{center}
\vspace{-0.4cm}
\caption{Results of fine-tuning on MNLI, SST-2, and two versions of SQuAD datasets. The two numbers of F1 and EM for each SQuAD dataset are first averaged. The FLOPs here means the floating-point operations in FFN layer or MoE layer. We only report the FLOPs in FFN or MoE layer because FLOPs at other layers are the same. We also compare the inference speed on TPU v3-8 to show the usability of dense model. The benefits here is the MoE benefits we proposed in Eq.~\ref{eq:moe_benefits}.}
\label{nlp:table:main}
%\vspace{-0.5cm}
\end{table*}

\subsubsection{Results on ImageNet}
\label{Experiments:CV:ResultsImageNet}

%\end{wraptable}

%The MoE benefits means the percentage of improvement from MoE can be preserved in the student dense model. We propose this metric to measure the ability to integrate knowledge from pre-trained MoE model.
We report the top-1 accuracy and MoE benefits on ImageNet in Table~\ref{cv:table:main}. In this table, as we defined in Eq.~\ref{eq:moe_benefits}, the MoE benefits means how much improvement the dense model preserved, after knowledge integration. First, after investigating four different KG methods, the SVD-based integration method performs best. Therefore, we set the SVD-based method as the default choice in the following experiments. Top-K-based integration method performs comparably with SVD-based method at large scale but slightly worse at base level. We suggest the reason is large model has larger capacity and is more robust to sparse column drop. Also, we observe that OneS-L-SVD achieves $78.4\%$ top-1 accuracy on ImageNet with only 15M parameters. Compared with the strongest Switch-L, our model has $0.6$ points improvement. Compared with the teacher model, OneS-L-SVD outperforms WideNet-B by $0.9\%$ with half of the parameters. As a final result, OneS-L-SVD achieves comparable performance with ViT-B with only $17\%$ trainable parameters. More importantly, in \cite{Xue2021GoWI}, without MoE, WideNet-L can obtain only achieve $76.9\%$ top-1 accuracy. Our OneS has the totally same architecture as that, but we can achieve $78.4\%$ accuracy. That is, our OneS-L-SVD preserves $61.7\%$ improvement (\ie MoE benefits) from WideNet. In addition, our OneS-B-SVD achieves $57.7$ MoE benefits, which outperforms the strongest baseline (\ie Switch) by $23.1$ points. Such results show the effectiveness of knowledge integration.

%Observe that WideNet-H achieves the best performance and significantly outperforms ViT and ViT-MoE models on ImageNet. Compared with the strongest baseline, our WideNet-H outperforms ViT-B by $1.5\%$ with less trainable parameters. Even if we use the smallest model, WideNet-B, it still achieves comparable performance with ViT-L and ViT-MoE-B with over $4 \times$ less trainable parameters. When we scale up to WideNet-L, it has surpassed all baselines with half trainable parameters of ViT-B and $0.13 \times$ parameters of ViT-L. 

%\end{wraptable}

\subsubsection{Results on Cifar10}
\label{Experiments:CV:ResultsCifar10}

We further fine-tune our dense student model, OneS on Cifar10 in this part. As shown in Table~\ref{cv:table:cifar}, our OneS-L still outperforms our baselines, Switch-B and Switch-L, by $0.3\%$ and $0.6\%$ respectively. The OneS-L can even achieve comparable performance with WideNet-B with $0.33 \times$ trainable parameters. OneS-B also achieves better performance than Switch-B due to knowledge gathering. In summary, the results on Cifar10 show the improvement of pre-training on ImageNet can propagate to the downstream task.

\subsection{Natural Language Processing}
\label{Experiments:NLP}

%\subsubsection{Experimental Settings}
%\label{Experiments:NLP:setting}

\textbf{Experimental settings} Similar to experiments on computer vision tasks, we still have two stages of training in natural language processing. The difference is, following existing works~\cite{lan2019albert,devlin-etal-2019-bert,Xue2021GoWI}, we focus on the performance of downstream tasks instead of pre-training. \textbf{Datasets} We use English Wikipedia~\cite{devlin-etal-2019-bert} and BOOKCORPUS~\cite{zhu2015aligning} as our pre-training corpus. For fine-tuning, we evaluate our work on General Language Understanding Evaluation (GLUE) benchmark~\cite{wang-etal-2018-glue}, two different versions of the Stanford Question Answering (SQuAD) dataset~\cite{rajpurkar-etal-2016-squad,rajpurkar-etal-2018-know}. For GLUE experiments, we report median over 5 runs following existing works~\cite{lan2019albert,Xue2021GoWI}. \textbf{Baselines} Similar to the experiments on computer vision, we still select Distill and Switch as our direct baselines, although our work is the first one focusing on this task. The student model here also has the same architecture as ALBERT except for the individual layer normalization~\cite{Xue2021GoWI}.  Therefore, another baseline is ALBERT. We expect our OneS can outperform ALBERT with the almost same architecture, a comparable number of parameters, and the same pre-training dataset. \textbf{Hyper-parameters} After initialization, we further train OneS by a linear combination of masked language modeling loss, sentence order prediction loss, and soft knowledge distillation loss. Following~\cite{sanh2019distilbert}, we only feed the logits of masked language modeling loss to $L_{distill}$. We still freeze all trainable weights of the teacher MoE model (WideNet) in the training stage of OneS. $\alpha$ is set as $0.75$, and $\lambda$ is $0.25$ in this part. The ablation study of these settings can be found in Appendix~\ref{Appendix:ablation-svd-ratio}. Other detailed hyper-parameters can be found in Appendix~\ref{sec:nlp-hyper}.

%datasets, baselines, hyper-parameters

\subsubsection{Results on NLU benchmarks}
\label{Experiments:NLP:ResultsNLU}

After pre-training, we fine-tune our OneS without distillation loss. Such a setting is different from existing work on distilling language models. The reason is, one of our goals is to obtain an easy-to-use model without expert routing. If we still have an MoE teacher, the downstream fine-tuning still requires complicated hardware and software co-design for MoE. The results on downstream natural language understanding tasks are shown in Table~\ref{nlp:table:main}. In general, we can observe OneS outperforms ALBERT and baselines (\ie Distill and Switch) on all tasks by achieving $88.2\%$ MoE benefits. For instance, on four tasks, OneS surpass Switch by $0.42$ on average. Also, we achieve $53.2\%$ and $51.7\%$ MoE benefits over Switch and Distill, respectively. On a few tasks, \eg SQuAD1.1 and SST-2, OneS can even outperform the teacher MoE model, WideNet. We suggest that MoE model tends to overfit on small datasets. OneS has MoE's knowledge but a dense structure, so that the benefits from pre-training can propagate to downstream tasks easier.

Compared with MoE model, another strength of our OneS is the inference speed. The reason why MoE is so slow is, MoE model has gating function and sparse einsum operators due to conditional computation, which would reduce the computational efficiency. However, our model can achieve $3.7 \times$ inference speedup. Please note that WideNet only uses $2.4 \times$ FLOPs at MoE layers. For other layers, WideNet has the same computation cost as OneS or ALBERT, so global FLOPs is less than $2.4$ times of OneS. Therefore, although one reason why OneS can achieve such high efficiency is less computation, another important reason is, the dense model is more hardware-friendly than sparse MoE model.

\subsection{Ablation study}
\label{Experiments:CV:Ablation}

%\begin{wraptable}{r}{0.5\textwidth}
\begin{table}[t]

%\vspace{-0.5cm}

\vspace{-0.3cm}
\begin{center}
\begin{small}
%\begin{sc}
\begin{tabular}{l  l}
\toprule
Model        & ImageNet \\ \midrule
OneS-B               &    75.7    \\
~w/o KG               &    73.8    \\
~w/o KD               &    75.0    \\
~w/o KG \& KD                &    72.8    \\ \midrule
OneS-L               &    78.4    \\
~w/o KG               &    77.3    \\
~w/o KD               &    77.6    \\
~w/o KG \& KD                &   76.9     \\
\bottomrule
\end{tabular}
\end{small}
%\end{sc}
\vspace{-0.3cm}
\end{center}
\caption{Top-1 Accuracy of ablation study on ImageNet to investigate the contributions of knowledge gathering (KG) and knowledge distillation (KD). The KG here is using SVD-KG, and the KD here is using soft-distillation, as we found they perform better by investigation.}
\label{cv:table:imagenet-ablation}
\vspace{-0.3cm}
\end{table}
%\end{wraptable}

%\begin{table}[t]

%\begin{figure}[t]

We conduct four sets of ablation studies in this work. The first set is to investigate the contributions of knowledge gathering and knowledge distillation. As shown in Table~\ref{cv:table:imagenet-ablation}, there is a significant performance drop without knowledge gathering, which shows the knowledge included in pre-trained sparse model is critical to improve the student model's performance. For the model without KD, in this experiment, we adopt the $L_{main}$ in Eq.~\ref{eq:loss} as the only loss function. We can see the knowledge distillation is helpful, as the prediction of teacher can instruct the student to mine knowledge in noisy weights gathered. In addition, when the dense model does not gather knowledge from MoE, the KD enables the training process of the lite model (\ie OneS-B) more stable. For the large model, removing both knowledge gathering and knowledge distillation will also harm the performance.

%\begin{wrapfigure}{r}{0.5\textwidth}
\begin{figure}[t]
%\vspace{-0.5cm}
\begin{center}
\centerline{\includegraphics[width=0.4\textwidth]{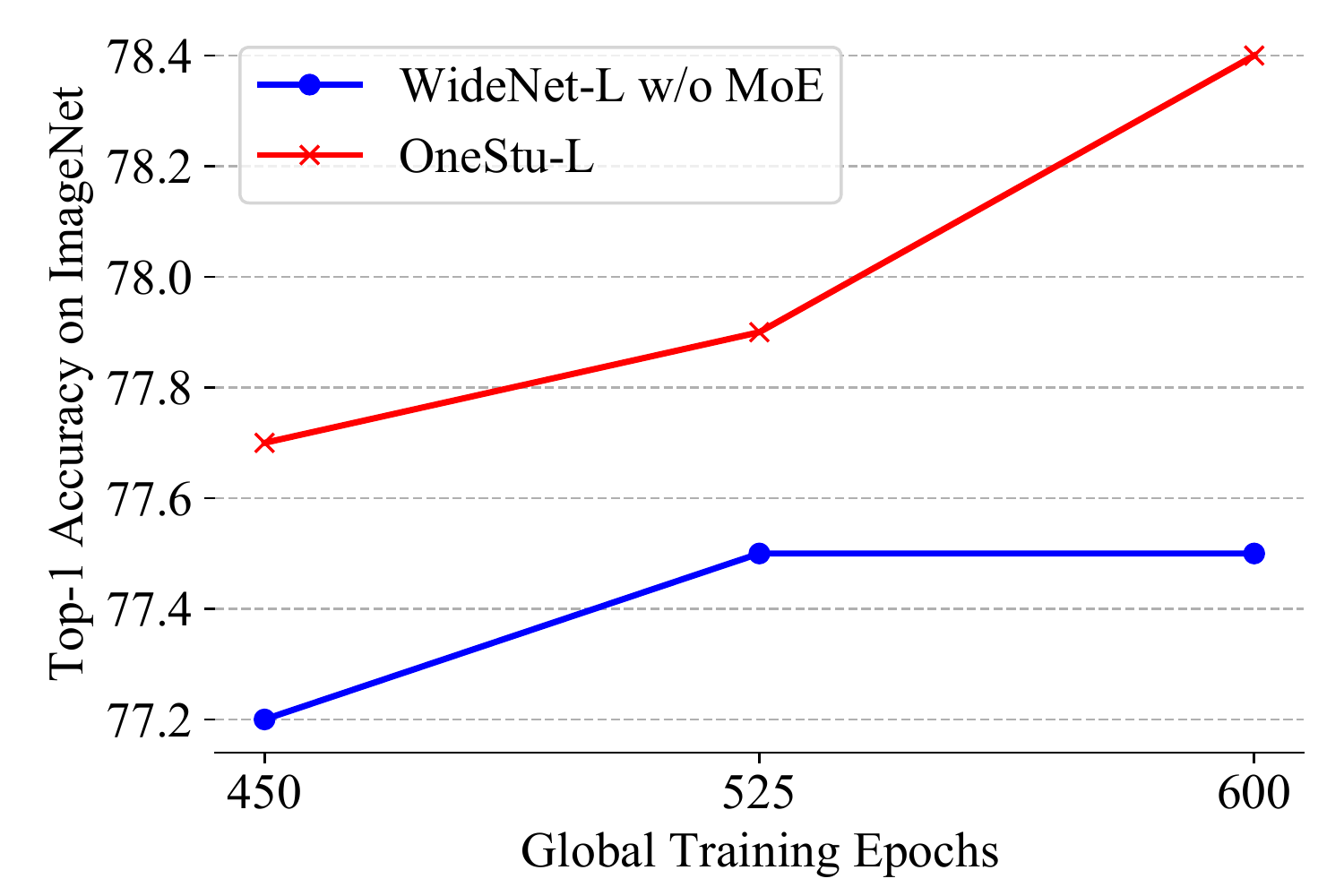}}
\caption{Top-1 Accuracy of ablation on ImageNet to investigate the contribution of more global training epochs.}
\label{fig:exp:Epochs}
\end{center}
\vspace{-0.9cm}
\end{figure}
%\end{wrapfigure}

Since we conduct two stages of training in our framework, the total training steps of OneS are more than the dense model trained from scratch without distillation. The second set of ablation study is to verify whether the improvement of our model is from more training iterations. To this end, we train the OneS without KG and KD from scratch for comparable global training epochs. We use OneS-L as a platform for this set of experiments because we observe the unstable training of OneS-B without both KG and KD. As shown in Figure~\ref{fig:exp:Epochs}, when training with comparable global epochs, our OneS outperforms baselines by a large margin consistently. Also, when scaling to more epochs, WideNet without MoE stops to improve, but our OneS can still obtain benefits from more training. We also investigate two types of knowledge distillation approaches, soft distillation~\cite{hinton2015distilling} and hard-label distillation~\cite{pmlr-v139-touvron21a}. The last set is to ablate the SVD ratio $\lambda$. Please see Appendix~\ref{Appendix:hard-label} and Appendix~\ref{Appendix:ablation-svd-ratio} for details.

%Hard-label distillation

%Same computation with training from scratch?

%\subsection{Number of Experts}
%\label{Experiments:NLP:NumExperts}

%\begin{table}[t]
%\caption{An investigation about how the number of experts impact the performance of students.}
%\label{nlp:table:num_experts}
%\vskip 0.15in
%\begin{center}
%\begin{sc}
%\begin{tabular}{l| l l l}
%\toprule
%\#Experts        & 4 & 8 & 16 \\ \midrule
%SQuAD1.1              & - & -  & -    \\
%SQuAD2.0              & - & -  & -    \\
%SST-2              & - & -  & -    \\
%MNLI              & - & -  & -    \\
%\bottomrule
%\end{tabular}
%\end{sc}
%\end{center}
%\vskip -0.1in
%\end{table}

%We investigate how the number of experts impact the performance on downstream tasks. No matter how many experts we have, we can only preserve one dense layer instead. Therefore, theoretically, it is challenging to keep the performance of student model improve along the number of experts in teacher. To investigate this, we conduct a set of experts to show how the performance of student models change when we use the teachers with different number of experts. As reported in \revise{Table}, 

%\subsubsection{SVD Analysis}
%\label{Experiments:NLP:SVDAnalysis}

%This is also our strong baseline.

\section{Conclusion and Future Work}
\label{Conclusion}

In this paper, inspired by the human education model, we propose knowledge integration, a new task to combine the effectiveness of the MoE model and the usability of dense model. As the first work focusing on this task, our solution is integrating knowledge in two steps (\ie knowledge gathering and knowledge distillation). Knowledge gathering focuses on gathering knowledge from pre-trained MoE to initialize dense student models. Knowledge distillation is to further refine the dense one. Experiments show that our OneS achieves outstanding effectiveness and efficiency on computer vision and natural language processing tasks. It is noteworthy our OneS can even preserve $88.2\%$ benefits from MoE with $0.42 \times$ FLOPs per MoE or FFN layer, $3.7 \times$ inference speedup, and $46\%$ trainable parameters.

In the future, we plan to explore more advanced knowledge gathering and distillation approaches to better integrate knowledge of MoE into a dense student. In addition, although most recent MoE-based transformer are using the same architecture for different experts, it is valuable to investigate the approach to gather knowledge from experts with different architectures. Last, we expect to adapt our approach to the extremely huge MoE model like GLaM~\cite{du2021glam}.

%%%%%%%%% REFERENCES
{\small
\bibliographystyle{ieee_fullname}
\bibliography{egbib}
}

\clearpage

\newpage

\appendix
\noindent\textbf{\Large Appendix}

\section{Knowledge Gathering Noise Analysis}\label{appendix:noise}

We are to discuss and analyze the induced noise during SVD knowledge gathering in this section. 

Given one MoE layer $\mathrm{MoE}(\cdot)$, the target of SVD-KG is to integrate its knowledge to a dense layer $g(\cdot)$ in the student model. For brevity, we set every expert and the dense student layer as the single linear layer. There are $E$ experts in MoE layer: $\{f^1,\dots, f^E \}$ with weights $\{W_f^1, \dots, W_f^E \}$ and bias $\{b_f^1, \dots, b_f^E \}$. The dense student layer is $g$ with weights $W_g$ and bias $b_g$. According to Eq.~\ref{eq:MoE}, the MoE layer can be written as:

\begin{equation}
\begin{aligned}
\mathrm{MoE}(x)&=\sum_{i=1}^E {G(x)}_i {e_i(x)}\\
&=\sum_{i=1}^E p_i h_i (W_f^i x+b_f^i)
\end{aligned}
\end{equation}
where $p$ is the routing score of router, $h$ is an index vector. For the selected experts, $h_i=1$, and $h_i=0$ for other unselected experts. Due to the load balance loss during MoE training, we can assume $p_i \approx 1.0 $ when $h_i=1$. Then, we can approximate MoE layer by SVD:

\begin{equation}
\begin{aligned}
\mathrm{MoE}(x) &= \sum_{i=1}^E p_i h_i (U_f^i S_f^i {V_f^i}^T x+b_f^i)  \\
&\approx \sum_{i=1}^E h_i ({U_f^i}_{K^i} {S_f^i}_{K^i} {V_f^i}_{K^i}^T x+b_f^i) \\
&\approx \sum_{i=1}^E h_i \sum_{j=1}^{K^i} {u_{f}^{ij}}_{K^i} {s_{f}^{ij}}_{K^i} {v_{f}^{ij}}_{K^i}^T x + \sum_{i=1}^E h_i b_f^i\\
\end{aligned}
\end{equation}

where $K^i$ is the selected rank of $i$-th expert.

According to Eq.~\ref{eq:assign_stu}, $g(\cdot)$ can be formulated as:

\begin{equation}
g(x)= \sum_{i=1}^E \sum_{j=1}^{K^i} {u_{f}^{ij}}_{K^i} {s_{f}^{ij}}_{K^i} {v_{f}^{ij}}_{K^i}^T x +\frac{1}{E} \sum_{i=1}^E b_f^i
\end{equation}
%where $K_g=\Sigma_{i=1}^E K^i$. 

For brevity, to analyze, we assume MoE layer here is to select the $1$-st expert, and then the MoE layer can be written as:

\begin{equation}
\mathrm{MoE}(x) \approx \sum_{j=1}^{K^1} {u_{f}^{1j}}_{K^1} {s_{f}^{1j}}_{K^1} {v_{f}^{1j}}_{K^1}^T x + b_f^1
\end{equation}

and the student dense layer:

\begin{equation}
\begin{aligned}
g(x)&=\sum_{j=1}^{K^1} {u_{f}^{1j}}_{K^1} {s_{f}^{1j}}_{K^1} {v_{f}^{1j}}_{K^1}^T x + b_f^1\\ 
&+ \sum_{i=2}^E \sum_{j=1}^{K^i} {u_{f}^{ij}}_{K^1} {s_{f}^{ij}}_{K^1} {v_{f}^{ij}}_{K^1}^T x \\
&+ \frac{1}{E} \sum_{i=2}^E b_f^i - \frac{E-1}{E} b_f^1
\end{aligned}
\end{equation}

Since the non-selected experts do not interact with the current input token $x$, we assume, for the non-selected experts, we let $\epsilon_1 = f^i(x)$ and $ \epsilon_1 \sim \mathcal{N}(\mu_1, \sigma_1^2)$ and $\epsilon_2 = b_f^i x$ and $ \epsilon_2 \sim \mathcal{N}(\mu_2, \sigma_2^2)$ According to Eq.~\ref{eq:auto_SVD}, $g(x)$ can be written as:

\begin{equation}
g(x)=\sum_{j=1}^{K^1} {u_{f}^{1j}}_{K^1} {s_{f}^{1j}}_{K^1} {v_{f}^{1j}}_{K^1}^T x + \lambda[(E-1)\epsilon_1 - \frac{E-1}{E} \epsilon_2]
\end{equation}

The low-rank approximation ensures $\sum_{j=1}^{K^1} {u_{f}^{1j}}_{K^1} {s_{f}^{1j}}_{K^1} {v_{f}^{1j}}_{K^1}^T + b_f^1$ cover most informative knowledge in the selected expert, and noise reduced linearly along $\lambda$. When we are integrating knowledge from experts, a smaller $\lambda$ is required to reduce noise.

\section{Hyper-parameters}

\subsection{Computer Vision}

\begin{table}[ht]
\caption{Hyper-parameters on ImageNet pre-training and Cifar10 finetuning. $\alpha$ and $\lambda$ are from Eq.~\ref{eq:loss} and Eq.~\ref{eq:auto_SVD}}
\label{tbl-hyper-parameter-pre-train}
\vskip 0.15in
\begin{center}
%\begin{sc}
\begin{tabular}{l l l}
\toprule
Parameter                  & ImageNet  & Cifar10   \\ \midrule
Epoch                     & 300     & 100     \\
Warmup Epochs             & 30    & 0       \\
Batch Size                & 4096  & 512      \\
Learning rate             & 0.004  & 0.03  \\
Weight Decay              & 0.1   & 0  \\
Dropout                   & 0.1   & 0.1    \\ 
Label smoothing           & 0.1    & 0     \\
Mixup prob.               & 0.5   & 0.5     \\
$\alpha$               & 0.25   & -     \\
$\lambda$               & 0.75   & -     \\
\bottomrule
\end{tabular}
%\end{sc}
\end{center}
\vskip -0.1in
\end{table}

Most hyper-parameters are set following existing works (\eg ViT, WideNet). The main difference is the learning rate. Since we are training from a dense model initialized by a MoE model. We observe that a large learning rate harms accuracy. We, therefore, set a smaller learning rate as 0.004 (0.01 in WideNet).

\subsection{Natural Language Processing}\label{sec:nlp-hyper}

%The pre-training hyper-parameters is shown in 

\begin{table}[ht]
\caption{Hyper-parameters on NLP downstream tasks fine-tuning.}
\label{tbl-hyper-parameter-nlp}
\vskip 0.15in
\begin{center}
%\begin{sc}
\begin{tabular}{l l l l}
\toprule
Parameter                  & SQuAD1.1/2.0 & MNLI & SST2  \\ \midrule
Steps                     & 3649/8144 & 10000 & 5234     \\
Warmup              & 365/814   & 1000 & 314     \\
Batch Size                & 48 & 128 & 128     \\
LR             & 5e-5/3e-5 & 3e-5 & 4e-5  \\
Dropout           & 0/0   & 0  & 0   \\
Max Length               & 384/512  & 512 & 512  \\
\bottomrule
\end{tabular}
%\end{sc}
\end{center}
\vskip -0.1in
\end{table}

We follow the hyper-parameters in \cite{devlin-etal-2019-bert,lan2019albert,Xue2021GoWI} and the final hyper-parameters are reported in Table~\ref{tbl-hyper-parameter-nlp}.

\section{Hard-label Distillation}\label{Appendix:hard-label}

\subsection{Method}

The hard-label distillation takes the hard decision of the teacher as a true label. In other words, it treats the knowledge distillation task as a typical classification task, supervised by both the prediction from the teacher and ground truth.

\begin{equation}
\mathrm{L}_{distill}^{hard} = \mathrm{L_{CE}}(\omega(z_{s}), \mathrm{argmax}(z_{t}))
\end{equation}
where $L_{CE}$ is the cross-entropy loss, $\mathrm{argmax}$ is used to obtain the hard label of teacher's prediction. 

\subsection{Evaluation}

\begin{table}[h]
\caption{Top-1 Accuracy of different knowledge distillation approaches On ImageNet.}
\label{cv:table:loss}
%\vskip 0.1in
\begin{center}
\begin{small}
%\begin{sc}
\begin{tabular}{l  l}
\toprule
Approach        & ImageNet \\ \midrule
Soft distillation               &    \textbf{75.7}    \\
Hard-label distillation         &    75.4    \\
\bottomrule
\end{tabular}
%\end{sc}
\end{small}
\end{center}
%\vskip -0.3in
\end{table}

We investigate two types of knowledge distillation approaches, soft distillation~\cite{hinton2015distilling} and hard-label distillation~\cite{pmlr-v139-touvron21a}, as we introduced in Section~\ref{sec:KD}. The results is reported in Table~\ref{cv:table:loss}. We observe that hard-label distillation can achieve comparable performance with soft distillation. Since soft distillation is popular in more tasks and has slightly better performance, we suggest using soft distillation as the default choice.

\section{Ablation Study on SVD Ratio}\label{Appendix:ablation-svd-ratio}

\begin{figure}[h]
%\begin{wrapfigure}{r}{0.5\textwidth}
%\vskip 0.1in
\begin{center}
\centerline{\includegraphics[width=0.4\textwidth]{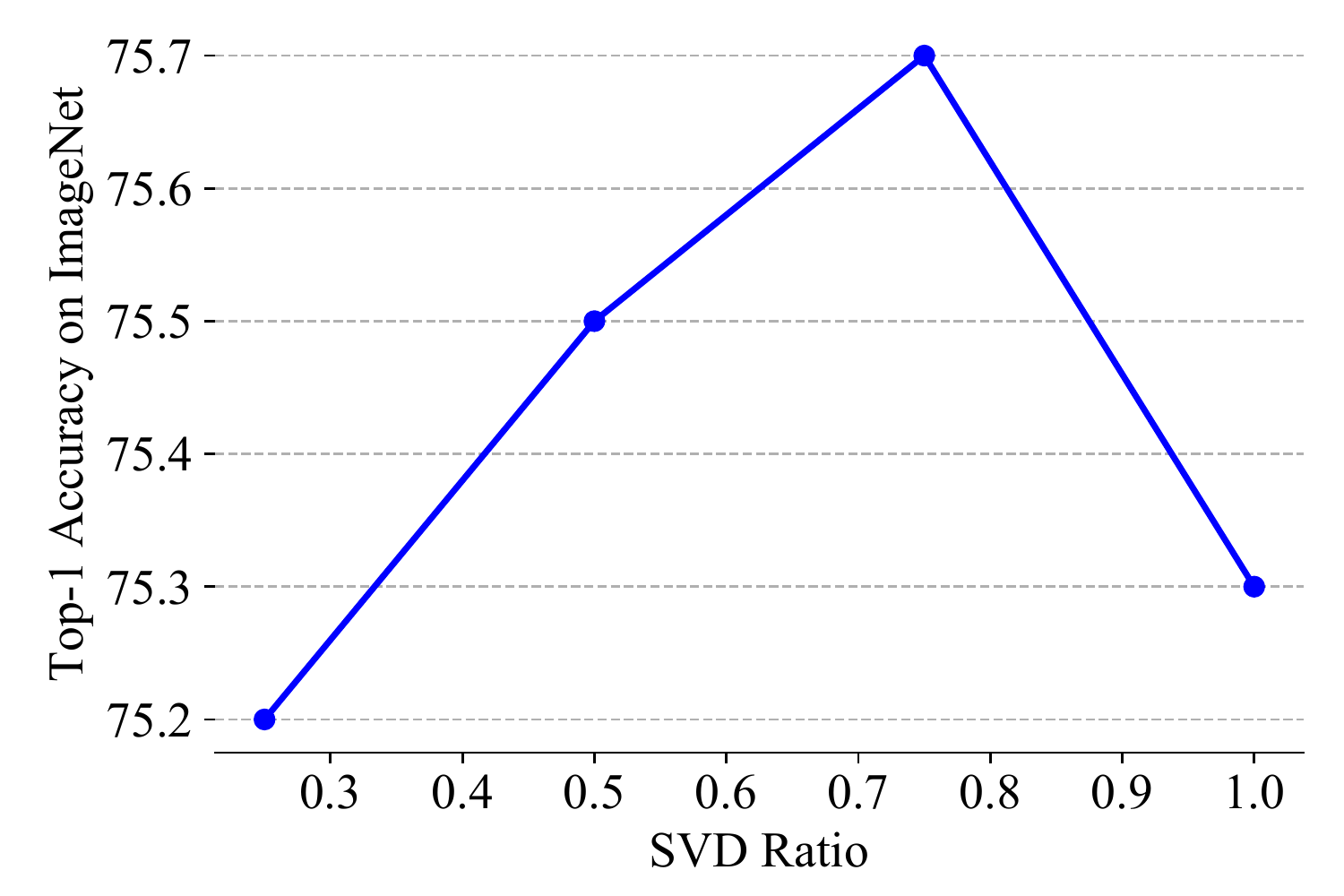}}
%\vskip -0.1in
\caption{Top-1 Accuracy of ablation on ImageNet to ablate the SVD ratio $\lambda$.}
\label{fig:exp:svd}
\end{center}
%\vskip -0.3in
\end{figure}
%\end{wrapfigure}

We also conduct ablation study on SVD ratio $\lambda$, which denotes the ratio of selected k. As shown in Figure~\ref{fig:exp:svd}, when $\lambda=0.75$, OneS-B achieves sweet point.

\section{Experimental Justification for Less Knowledge in Bias}\label{Appendix:less-know-bias}

\begin{table}[h]
\caption{Top-1 Accuracy of MoE model without bias.}
\label{cv:table:bias}
%\vskip 0.1in
\begin{center}
\begin{small}
%\begin{sc}
\begin{tabular}{l  l}
\toprule
Approach        & ImageNet \\ \midrule
WideNet-B              &    77.5    \\
WideNet-B w/o bias        &    77.3    \\
\bottomrule
\end{tabular}
%\end{sc}
\end{small}
\end{center}
%\vskip -0.3in
\end{table}

We re-trained the teacher MoE model (\ie WideNet-B) without bias in MoE layer. As shown in Table~\ref{cv:table:bias}, we found that there is no obvious performance drop. That is, the bias in MoE layer has little impact on results, which means there is less knowledge than weights.

\section{Related Work}

\subsection{Mixture-of-Experts}

MoE has shown promising results on various tasks. Recent works scaled a dense model to a sparse one by MoE. Faster convergence speed of MoE can save the global computation cost. One typical way to use MoE is, by replacing the FFN layer in transformer~\cite{vaswani2017attention} by an MoE layer. Lepikhin \etal~\cite{lepikhin2020gshard} first scale machine translation transformer model to 600 million parameters using automatic sharding. After that, Fedus \etal~\cite{fedus2021switch} further scales the transformer to trillion parameter models with simple and efficient sparsity and shows promising results on natural language understanding. In computer vision, ViT-MoE~\cite{ruiz2021scaling} matches SoTA performance on ImageNet using $14.7$ billion of parameters, while requiring as little as half of the computation at inference time. Recent work~\cite{lou2021sparse} investigated the MoE usage on MLP-Mixer, which also achieved better effectiveness and efficiency than the dense model. Instead of scaling up, this work uses and fixes the pre-trained MoE model. The core target is to combine the effectiveness of MoE and the usability of dense model.

\subsection{Knowledge Integration}

Knowledge inheritance~\cite{qin2021knowledge} is related to our knowledge integration. Knowledge inheritance usually inherits knowledge from small pre-trained model and then speed-up the training of large models. Contrastively, our work is integrating knowledge from a large MoE model. Sun \etal~\cite{sun2019ernie} proposed to integrate knowledge by using knowledge masking strategies. Please note our knowledge integration is different from theirs. Instead of a self-supervised learning approach to integrate knowledge from data, our work is to integrate knowledge from pre-trained MoE. There are also a few works focusing on inheriting knowledge from a dense model to initialize a MoE model, which can be seen as an inverse process of ours. For instance, Zhang \etal~\cite{zhang2022cpm} duplicated dense model multiple times to initialize MoE models.  Zhang \etal~\cite{zhang2021moefication} proposed MoEfication. The proposed approach is to inherit knowledge from a dense model and obtain an MoE model with comparable parameters to reduce the computation cost. In general, MoEfication is a sparsification approach. In Switch Transformer~\cite{fedus2021switch}, authors tried to initialize trainable parameters except for MoE layers to speed-up MoE training, although their main purpose is to scale transformer to trillions of parameters.

\end{document}